\documentclass{article}

\usepackage[preprint,nonatbib]{neurips_2022}

\usepackage[utf8]{inputenc} % allow utf-8 input
\usepackage[T1]{fontenc}    % use 8-bit T1 fonts
\usepackage{url}            % simple URL typesetting
\usepackage{booktabs}       % professional-quality tables
\usepackage{amsfonts}       % blackboard math symbols
\usepackage{nicefrac}       % compact symbols for 1/2, etc.
\usepackage{microtype}      % microtypography
\usepackage[dvipsnames]{xcolor}

\usepackage{amsmath}
\usepackage{amssymb}
\usepackage{amsthm}
\usepackage{mathrsfs}
\usepackage{dsfont}

\usepackage{algorithm}
\usepackage{algpseudocode}

\DeclareMathOperator{\MS}{MS}
\DeclareMathOperator{\diag}{diag}
\DeclareMathOperator{\e}{e}

\RequirePackage[colorlinks,citecolor=blue,urlcolor=blue]{hyperref}

\usepackage{tikz}
\usepackage{pgfplots}
\usepgfplotslibrary{statistics}
\usepgfplotslibrary{groupplots}
\pgfplotsset{compat=newest}
\usetikzlibrary{arrows,arrows.meta,calc,quotes,angles,shapes,plotmarks}

\title{Langevin algorithms for very deep Neural Networks with application to image classification}

\author{%
  Pierre ~Bras\thanks{\url{https://www.lpsm.paris/pageperso/brasp/}}\\
  Laboratoire de Probabilit\'es, Statistique et Mod\'elisation\\
  Sorbonne Universit\'e\\
  \texttt{pierre.bras@sorbonne-universite.fr}\\
}

\begin{document}

\maketitle

\begin{abstract}
Training a very deep neural network is a challenging task, as the deeper a neural network is, the more non-linear it is.
We compare the performances of various preconditioned Langevin algorithms with their non-Langevin counterparts for the training of neural networks of increasing depth. For shallow neural networks, Langevin algorithms do not lead to any improvement, however the deeper the network is and the greater are the gains provided by Langevin algorithms. Adding noise to the gradient descent allows to escape from local traps, which are more frequent for very deep neural networks.
%We also make similar observations for highway networks, which were introduced to tackle the difficulty of training very deep networks.
Following this heuristic we introduce a new Langevin algorithm called Layer Langevin, which consists in adding Langevin noise only to the weights associated to the deepest layers.
We then prove the benefits of Langevin and Layer Langevin algorithms for the training of popular deep residual architectures for image classification.
\end{abstract}

\section{Introduction}
Langevin algorithms are widely used for the training of neural networks in a Bayesian setting \cite{welling2011, jospin2020}. Adding a small exogenous noise adds regularization to the training and allows to quantify the degree of uncertainty on the parameters.
In this paper, we consider Langevin algorithms directly used for stochastic optimization of neural networks in a non-Bayesian setting and compare their performances with non-Langevin stochastic gradient algorithms.
As it was noted in \cite{neelakantan2015, asgld}, adding gradient noise can in fact improve the learning. Similarly, noisy activation functions \cite{gulcehre2016, shridhar2019} may yield better learning for very deep neural networks. Indeed, the noise provides regularization and allows to escape from traps for the gradient descent such as local minima and saddle points \cite{dauphin2014}.
Moreover, the deeper the neural network is, the more non-linear it is, thus increasing the number of such traps.
Non-convex optimization through Langevin algorithms shares heuristics with simulated annealing which consists in sampling with respect to a Gibbs measure where the noise parameter gradually decreases to zero \cite{laarhoven1987, bras2021-2}.

Many advances in supervised learning were made possible using very deep neural networks, which are able to tackle much more difficult problems than shallow ones \cite{krizhevsky2012, montufar2014, lecun2015}, in particular as it comes to image classification \cite{szegedy2015, simonyan2015, residual, huang2017}.
Still, deep neural networks which consist in a succession of dense layers are considerably more difficult to train \cite{glorot2010, dauphin2014} and may run into vanishing gradient problems \cite{hochreiter1991, hanin2018}. Without proper adaptation or training, they show poor performance.
To cope with this issue, highway networks \cite{highway} and residual networks \cite{residual} were introduced. Their many successive layers behaves either as a dense layer or as the identity function, allowing the gradient information to propagate trough the successive layers.

We compare the benefits of preconditioned Langevin algorithms \cite{li2015} for various architectures and depths of neural networks and we proceed to side-to-side comparison of Langevin algorithms with their respective non-Langevin counterparts. The purpose of our experiments is to compare different methods on the same model architecture, not to achieve state-of-the-art results.
For shallow networks, there is no benefit in using Langevin algorithms as it only adds noise to the gradient descent and brings a less accurate estimation of the minimum.
However, we observe that the deeper the network is, the greater are the gains provided by Langevin algorithms.
%We make similar observations for highway networks, however they are valid only from a much larger depth.
%These observations are independent of the dimension (number of parameters) of the network: Langevin algorithms perform better on deeper networks but not necessarily on very high-dimensional but shallow ones.

%We then give a further heuristic to explain the better performances of Langevin algorithms in this context.
Since the most important non-linearities of the network are contained in the deepest layers, we introduce a new optimization method that we call Layer Langevin algorithm, which consists in training the network by adding Langevin noise only to the training of some layers and not to the other layers. In particular, we choose the Langevin layers to be the $k$ first (deepest) layers for some integer $k$.
%We observe that choosing the Langevin layers to be the $k$th first layers for some integer $k$ may lead to better performances, whereas choosing the Langevin layers to be the $k$th last layers yields poor performances, thus confirming the fact that the deepest layers bear the largest non-linearities and are more subject to Langevin optimization.
We then highlight the possibilities of training acceleration using Langevin and Layer Langevin methods on deep residual networks \cite{residual}
%and dense convolutional networks \cite{huang2017}
for image classification.

Our code for the numerical experiments is available at \url{https://github.com/Bras-P/deep-layer-langevin}. It includes in particular ready-to-use Langevin optimizers and Layer Langevin optimizers as instances of the TensorFlow \texttt{Optimizer} base class and a demonstration notebook.

\section{Very deep neural networks}

Training of very deep neural networks is a significantly more challenging task than for shallow networks \cite{glorot2010, dauphin2014}.
Let us write the output of a neural network with $K$ layers and with weights $\theta=(\theta^1, \ldots, \theta^K)$ as
\begin{equation}
\psi_\theta(x) = \varphi^K_{\theta^K} \circ \cdots \circ \varphi^1_{\theta^1}(x),
\end{equation}
where $\varphi^1, \ldots, \varphi^K$ are activation function and where $\varphi^k_{\theta^k} : x \mapsto \varphi^k(\theta^k \cdot x)$ at every unit.
%We omit the biases for simplicity.
Denoting
\begin{equation}
\Phi_k(x) := \varphi^k_{\theta^k} \circ \cdots \circ \varphi^1_{\theta^1} (x)
\end{equation}
for $1 \le k \le K$ and $\Phi_0(x) := x$, then the gradient reads for $1 \le k \le K$:
\begin{equation}
\label{eq:composed_derivative}
\nabla_{\theta^k} \psi_\theta (x) = \left(\nabla_{\theta^K} \varphi_{\theta^K} \circ \Phi_{K-1}(x) \right) \cdot \cdots \cdot \left( \nabla_{\theta^k} \varphi_{\theta^k} \circ \Phi_{k-1}(x) \right).
\end{equation}
Thus heuristically, the deeper the layer is, the more the gradient with respect to the parameters of this layer has annealing points, since more factors appear in \eqref{eq:composed_derivative}, hinting that deep layers show more non-linearities and local traps.

% \nabla_{\theta_{K-k}} \psi_{\theta}(x) = \prod_{j=1}^{k+1} \left( (\varphi^j)'_{\theta_j} \circ \Phi_{j-1}(x) \right) \cdot \Phi_{K-k-1}(x).

\section{Langevin algorithms for the training of deep neural networks}
\label{sec:experiments1}

\subsection{Experimental setting}

In our experiments we use the following datasets.
The MNIST dataset \cite{mnist} is composed of $28 \times 28$ grayscale images of handwritten digits (from 0 to 9). 60.000 images are used for training and 10.000 images are used for test.
The CIFAR-10 and the CIFAR-100 datasets \cite{cifar} consist in RGB images of size $32 \times 32$ belonging to 10 and 100 different classes respectively. For both datasets 50.000 images are used for training and 10.000 images are used for test.
%We implement our algorithms using python and tensorflow librairies.

The neural networks are trained using preconditioned Langevin algorithms with per-dimension adaptive stepsize \cite{li2015} with different choices of preconditioner. That is, for a preconditioner rule $(P_n)$ the Langevin update reads
\begin{align}
& g_{n+1} = \nabla_\theta V(\theta_n;\mathcal{D}_{n+1}) \\
\label{eq:preconditioned_langevin}
& \theta_{n+1} = \theta_n - \gamma_{n+1} P_{n+1} \cdot g_{n+1} + \sigma \sqrt{\gamma_{n+1}} \mathcal{N}(0, P_{n+1}), 
\end{align}
where $\sigma \in (0,\infty)$ controls the amount of injected noise, $(\gamma_n)$ is the non-increasing learning rate sequence, $V$ denotes the objective function and where $\nabla_\theta V(\theta_n;\mathcal{D}_n)$ stands for the mean gradient computed on a subset $\mathcal{D}_n$ of the dataset. The corresponding preconditioned non-Langevin algorithm follows the same update as in \eqref{eq:preconditioned_langevin} without Gaussian noise. In our experiments we use the RMSprop \cite{duchi2011, li2015}, the Adam \cite{adam} and the Adadelta \cite{adadelta} preconditioners and we call the Langevin version of these algorithms as L-RMSprop, L-Adam and L-Adadelta respectively.
The preconditioner rules are given in Algorithms \ref{algo:rms_prop}, \ref{algo:adam}, \ref{algo:adadelta} respectively. Note that depending on the algorithm version, in the update \eqref{eq:preconditioned_langevin} the gradient $g_{n+1}$ can be replaced by an averaged gradient over the past iterations as this in the case in Adam (Algorithm \ref{algo:adam}) i.e. momentum gradient is used.
While comparing some preconditioned method with its Langevin counterpart, we ensure that both training procedures start with the same initial weights.

\begin{minipage}{0.46\textwidth}
\begin{algorithm}[H]
\caption{RMSprop update}\label{algo:rms_prop}
\begin{algorithmic}
\State \textbf{Parameters:} $\alpha, \lambda > 0$
\State $\MS_{n+1} = \alpha \MS_n + (1-\alpha) g_{n+1} \odot g_{n+1}$
\State $P_{n+1} = \diag\left( \mathds{1} \oslash \left(\lambda \mathds{1} + \sqrt{\MS_{n+1}}\right) \right)$
\State $\theta_{n+1} = \theta_n - \gamma_{n+1} P_{n+1} \cdot g_{n+1}$
\end{algorithmic}
\end{algorithm}
\end{minipage}
\hfill
\begin{minipage}{0.46\textwidth}
\begin{algorithm}[H]
\caption{Adam update}\label{algo:adam}
\begin{algorithmic}
\State \textbf{Parameters:} $\beta_1, \beta_2, \lambda > 0$
\State $M_{n+1} = \beta_1 M_n + (1-\beta_1) g_{n+1} $
\State $\MS_{n+1} = \beta_2 \MS_n + (1-\beta_2) g_{n+1} \odot g_{n+1} $
\State $\widehat{M}_{n+1} = M_{n+1}/(1-\beta_1^{n+1}) $
\State $\widehat{\MS}_{n+1} = \MS_{n+1}/(1-\beta_2^{n+1}) $
\State $P_{n+1} = \diag\big( \mathds{1} \oslash \big(\lambda \mathds{1} + \sqrt{\widehat{\MS}_{n+1}}\big) \big) $
\State $\theta_{n+1} = \theta_n - \gamma_{n+1} P_{n+1} \cdot \widehat{M}_{n+1} .$
\end{algorithmic}
\end{algorithm}
\end{minipage}
~
\begin{algorithm}
\caption{Adadelta update}\label{algo:adadelta}
\begin{algorithmic}
\State \textbf{Parameters:} $\beta_1, \beta_2, \lambda > 0$
\State $\MS_{n+1} = \beta_1 \MS_n + (1-\beta_1) g_{n+1} \odot g_{n+1} $
\State $P_{n+1} = \diag\big( \widehat{MS}_n + \lambda \mathds{1} \oslash \big(\lambda \mathds{1} + \sqrt{\widehat{\MS}_{n}}\big) \big) $
\State $\theta_{n+1} = \theta_n - \gamma_{n+1} P_{n+1} \cdot g_{n+1}.$
\State $\widehat{MS}_{n+1} = \beta_2 \MS_n + (1-\beta_2) (\theta_{n+1}-\theta_n) \odot (\theta_{n+1}-\theta_n)$.
\end{algorithmic}
\end{algorithm}

\subsection{Plain and convolutional networks}
\label{sec:experiments_plain}

\begin{figure}
\centering

\begin{tikzpicture}
	\begin{groupplot}[group style={
					group size=2 by 4,
					horizontal sep=2ex,vertical sep=4ex,
					x descriptions at=edge bottom,
					group name=plots,
					vertical sep=0.15cm,
					},
			  ytick pos=left,
			  height=0.16\textheight,width=0.48\textwidth,
			  xmin=0, xmax=15.3, xlabel=Epochs,
			  grid=both,minor grid style={gray!25},major grid style={gray!25},
			  tick label style={font=\small},
			  ]
  		\nextgroupplot[legend to name=grouplegend1,legend columns=2, title=Test accuracy, title style={yshift=-1ex}, line width=1pt, mark size=2pt,ymin=0.9, ylabel=\textbf{(a)}, legend style={ font=\footnotesize}]
			\addplot[color=black, mark=*] %
				table[x=time,y=f,col sep=comma]{./data/dense/rmsprop_3x64/0_v_acc.csv};
			\addlegendentry[color=black]{RMSprop};
			\addplot[color=blue, mark=*, style=densely dotted] %
				table[x=time,y=f,col sep=comma]{./data/dense/rmsprop_3x64/1_v_acc.csv};
			\addlegendentry[color=black]{L-RMSprop};
			\addplot[color=red, mark=diamond*, style=densely dashed] %
				table[x=time,y=f,col sep=comma]{./data/dense/adam_3x64/0_v_acc.csv};
			\addlegendentry[color=black]{Adam};
			\addplot[color=olive, mark=square*, style=densely dashdotted] %
				table[x=time,y=f,col sep=comma]{./data/dense/adam_3x64/1_v_acc.csv};
			\addlegendentry[color=black]{L-Adam};
%			\addplot[color=Brown, mark=triangle*,]
%			\addplot[color=Plum, mark=pentagon*, style=densely dashdotted] 
		%
		%
  		\nextgroupplot[title=Train loss, title style={yshift=-1ex}, line width=1pt, line width=1pt,mark size=2pt,ymax=0.2,ytick pos=right,]
			\addplot[color=black, mark=*] %
				table[x=time,y=f,col sep=comma]{./data/dense/rmsprop_3x64/0_loss.csv};
			\addplot[color=blue, mark=*, style=densely dotted] %
				table[x=time,y=f,col sep=comma]{./data/dense/rmsprop_3x64/1_loss.csv};
			\addplot[color=red, mark=diamond*, style=densely dashed] %
				table[x=time,y=f,col sep=comma]{./data/dense/adam_3x64/0_loss.csv};
			\addplot[color=olive, mark=square*, style=densely dashdotted] %
				table[x=time,y=f,col sep=comma]{./data/dense/adam_3x64/1_loss.csv};
  		\nextgroupplot[line width=1pt,mark size=2pt, ymin=0.7, ylabel=\textbf{(b)}]
	  		\addplot[color=black, mark=*] %
		table[x=time,y=f,col sep=comma]{./data/dense/rmsprop_20x64/0_v_acc.csv};
			\addplot[color=blue, mark=*, style=densely dotted] %
				table[x=time,y=f,col sep=comma]{./data/dense/rmsprop_20x64/1_v_acc.csv};
			\addplot[color=red, mark=diamond*, style=densely dashed] %
				table[x=time,y=f,col sep=comma]{./data/dense/adam_20x64/0_v_acc.csv};
			\addplot[color=olive, mark=square*, style=densely dashdotted] %
				table[x=time,y=f,col sep=comma]{./data/dense/adam_20x64/1_v_acc.csv};
  		\nextgroupplot[line width=1pt,mark size=2pt,ymax=1, ytick pos=right,]
			\addplot[color=black, mark=*] %
				table[x=time,y=f,col sep=comma]{./data/dense/rmsprop_20x64/0_loss.csv};
			\addplot[color=blue, mark=*, style=densely dotted] %
				table[x=time,y=f,col sep=comma]{./data/dense/rmsprop_20x64/1_loss.csv};
			\addplot[color=red, mark=diamond*, style=densely dashed] %
				table[x=time,y=f,col sep=comma]{./data/dense/adam_20x64/0_loss.csv};
			\addplot[color=olive, mark=square*, style=densely dashdotted] %
				table[x=time,y=f,col sep=comma]{./data/dense/adam_20x64/1_loss.csv};
  	  	\nextgroupplot[line width=1pt,mark size=2pt, ymin=0.4, ylabel=\textbf{(c)}]
	  		\addplot[color=black, mark=*] %
		table[x=time,y=f,col sep=comma]{./data/dense/rmsprop_30x64/0_v_acc.csv};
			\addplot[color=blue, mark=*, style=densely dotted] %
				table[x=time,y=f,col sep=comma]{./data/dense/rmsprop_30x64/1_v_acc.csv};
			\addplot[color=red, mark=diamond*, style=densely dashed] %
				table[x=time,y=f,col sep=comma]{./data/dense/adam_30x64/0_v_acc.csv};
			\addplot[color=olive, mark=square*, style=densely dashdotted] %
				table[x=time,y=f,col sep=comma]{./data/dense/adam_30x64/1_v_acc.csv};
  		\nextgroupplot[line width=1pt,mark size=2pt,ymax=2., ytick pos=right,]
			\addplot[color=black, mark=*] %
				table[x=time,y=f,col sep=comma]{./data/dense/rmsprop_30x64/0_loss.csv};
			\addplot[color=blue, mark=*, style=densely dotted] %
				table[x=time,y=f,col sep=comma]{./data/dense/rmsprop_30x64/1_loss.csv};
			\addplot[color=red, mark=diamond*, style=densely dashed] %
				table[x=time,y=f,col sep=comma]{./data/dense/adam_30x64/0_loss.csv};
			\addplot[color=olive, mark=square*, style=densely dashdotted] %
				table[x=time,y=f,col sep=comma]{./data/dense/adam_30x64/1_loss.csv};
  		\nextgroupplot[line width=1pt,mark size=2pt, ylabel=\textbf{(d)}]
	  		\addplot[color=black, mark=*] %
		table[x=time,y=f,col sep=comma]{./data/dense/rmsprop_40x64/0_v_acc.csv};
			\addplot[color=blue, mark=*, style=densely dotted] %
				table[x=time,y=f,col sep=comma]{./data/dense/rmsprop_40x64/1_v_acc.csv};
			\addplot[color=red, mark=diamond*, style=densely dashed] %
				table[x=time,y=f,col sep=comma]{./data/dense/adam_40x64/0_v_acc.csv};
			\addplot[color=olive, mark=square*, style=densely dashdotted] %
				table[x=time,y=f,col sep=comma]{./data/dense/adam_40x64/1_v_acc.csv};
  		\nextgroupplot[line width=1pt,mark size=2pt, ytick pos=right,]
			\addplot[color=black, mark=*] %
				table[x=time,y=f,col sep=comma]{./data/dense/rmsprop_40x64/0_loss.csv};
			\addplot[color=blue, mark=*, style=densely dotted] %
				table[x=time,y=f,col sep=comma]{./data/dense/rmsprop_40x64/1_loss.csv};
			\addplot[color=red, mark=diamond*, style=densely dashed] %
				table[x=time,y=f,col sep=comma]{./data/dense/adam_40x64/0_loss.csv};
			\addplot[color=olive, mark=square*, style=densely dashdotted] %
				table[x=time,y=f,col sep=comma]{./data/dense/adam_40x64/1_loss.csv};

	\end{groupplot}
  	\node at (plots c1r1.north east) [inner sep=0pt,anchor=north, yshift=10ex] {\ref{grouplegend1}};
  	%\node[rotate=90] at (plots c1r1.south west) [yshift=7ex] {Test accuracy};
  	%\node[rotate=-90] at (plots c1r1.south east) [yshift=54ex] {Train loss};
\end{tikzpicture}

\caption{Training of neural networks of various depths on the MNIST dataset using Langevin algorithms compared with their non-langevin counterparts. (a): 3 hidden layers, (b): 20 hidden layers, (c): 30 hidden layers, (d): 40 hidden layers. The batch size is 512. The schedules are $\gamma_n=1\e-3$ and $\sigma=5\e-4$ for epochs 1 to 12 and $\gamma_n=1e-4$ and $\sigma=0$ beyond.}
\label{fig:dense:1}
\end{figure}
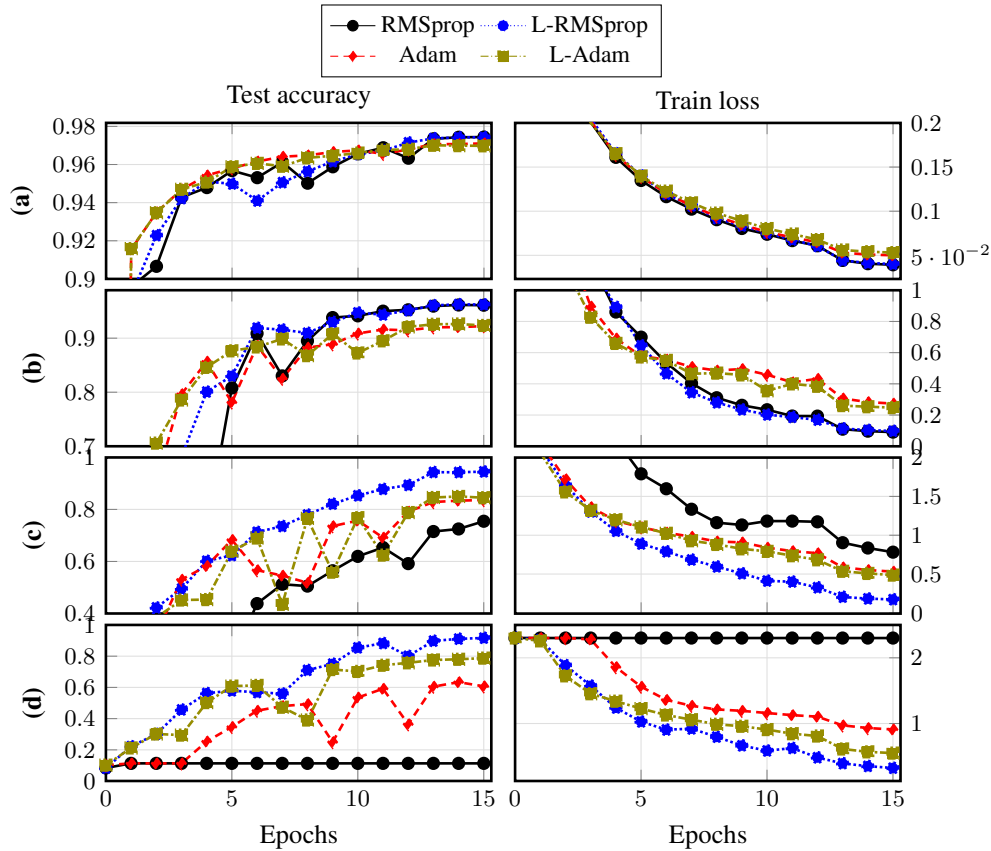

We first train fully connected feedforward neural networks on the MNIST dataset. The networks are composed of 3, 20, 30 and 40 hidden dense layers respectively with 64 units each and with ReLU activation, followed by one dense output layer.
The results are given in Figure \ref{fig:dense:1}.
We observe that for shallow neural networks, Langevin algorithms do not outperform their respective non-Langevin counterparts; they add noise to the gradient descent thus giving a less accurate estimate of the minimum value. In particular and as noted in the footnote in \cite{marceau2017}, we could not reproduce the good results from \cite{li2015} for plain networks with two hidden layers.
However, the deeper the network is, the greater the gains induced by Langevin algorithms compared with their respective non-Langevin counterparts are. We also display the value of the loss function on the training set in order to highlight that the better performances of the Langevin algorithms are not due to some overfitting effect.
Langevin algorithms indeed show improvements on 20-layer deep networks; beyond 30-layer deep networks, the gains are significant. The training of 40-layer deep networks with non-Langevin algorithms may run into the vanishing gradient problem, whereas such problem is avoided by Langevin algorithms. In the latter case of very deep training, preconditioned Langevin algorithms not only add noise preventing the vanishing of the gradient, they also help starting up the training in the right directions.
To obtain better results with Langevin algorithms, we recommend using a small coefficient $\sigma$, empirically ranging from $1\e-3$ to $5\e-5$.

\begin{figure}
\centering

\begin{tikzpicture}
	\begin{groupplot}[group style={
					group size=2 by 2,
					horizontal sep=2ex,vertical sep=4ex,
					x descriptions at=edge bottom,
					group name=plots,
					vertical sep=0.15cm,
					},
			  ytick pos=left,
			  height=0.16\textheight,width=0.48\textwidth,
			  xmin=0, xmax=20.3, xlabel=Epochs,
			  grid=both,minor grid style={gray!25},major grid style={gray!25},
			  tick label style={font=\small},
			  ]

  		\nextgroupplot[legend to name=grouplegend2,legend columns=2, title=Test accuracy, title style={yshift=-1ex}, line width=1pt, mark size=2pt,ymin=0.1, ylabel=\textbf{(a)}, legend style={ font=\footnotesize}]
			\addplot[color=black, mark=*] %
				table[x=time,y=f,col sep=comma]{./data/conv_dense/rmsprop_10x64/0_v_acc.csv};
			\addlegendentry[color=black]{RMSprop};
			\addplot[color=blue, mark=*, style=densely dotted]
				table[x=time,y=f,col sep=comma]{./data/conv_dense/rmsprop_10x64/1_v_acc.csv};
			\addlegendentry[color=black]{L-RMSprop};
			\addplot[color=red, mark=diamond*, style=densely dashed]
				table[x=time,y=f,col sep=comma]{./data/conv_dense/adam_10x64/0_v_acc.csv};
			\addlegendentry[color=black]{Adam};
			\addplot[color=olive, mark=square*, style=densely dashdotted]
				table[x=time,y=f,col sep=comma]{./data/conv_dense/adam_10x64/1_v_acc.csv};
			\addlegendentry[color=black]{L-Adam};
%			\addplot[color=Brown, mark=triangle*,]
%			\addplot[color=Plum, mark=pentagon*, style=densely dashdotted] 

  		\nextgroupplot[title=Train loss, title style={yshift=-1ex}, line width=1pt, line width=1pt,mark size=2pt,ymax=2.,ytick pos=right,]
			\addplot[color=black, mark=*] %
				table[x=time,y=f,col sep=comma]{./data/conv_dense/rmsprop_10x64/0_loss.csv};
			\addplot[color=blue, mark=*, style=densely dotted]
				table[x=time,y=f,col sep=comma]{./data/conv_dense/rmsprop_10x64/1_loss.csv};
			\addplot[color=red, mark=diamond*, style=densely dashed]
				table[x=time,y=f,col sep=comma]{./data/conv_dense/adam_10x64/0_loss.csv};
			\addplot[color=olive, mark=square*, style=densely dashdotted] %
				table[x=time,y=f,col sep=comma]{./data/conv_dense/adam_10x64/1_loss.csv};

  		\nextgroupplot[line width=1pt,mark size=2pt, ylabel=\textbf{(b)}]
	  		\addplot[color=black, mark=*] %
		table[x=time,y=f,col sep=comma]{./data/conv_dense/rmsprop_30x64/0_v_acc.csv};
			\addplot[color=blue, mark=*, style=densely dotted]
				table[x=time,y=f,col sep=comma]{./data/conv_dense/rmsprop_30x64/1_v_acc.csv};
			\addplot[color=red, mark=diamond*, style=densely dashed]
				table[x=time,y=f,col sep=comma]{./data/conv_dense/adam_30x64/0_v_acc.csv};
			\addplot[color=olive, mark=square*, style=densely dashdotted]
				table[x=time,y=f,col sep=comma]{./data/conv_dense/adam_30x64/1_v_acc.csv};

  		\nextgroupplot[line width=1pt,mark size=2pt,ymax=2.3, ytick pos=right,]
			\addplot[color=black, mark=*]
				table[x=time,y=f,col sep=comma]{./data/conv_dense/rmsprop_30x64/0_loss.csv};
			\addplot[color=blue, mark=*, style=densely dotted]
				table[x=time,y=f,col sep=comma]{./data/conv_dense/rmsprop_30x64/1_loss.csv};
			\addplot[color=red, mark=diamond*, style=densely dashed]
				table[x=time,y=f,col sep=comma]{./data/conv_dense/adam_30x64/0_loss.csv};
			\addplot[color=olive, mark=square*, style=densely dashdotted]
				table[x=time,y=f,col sep=comma]{./data/conv_dense/adam_30x64/1_loss.csv};
  		
	\end{groupplot}
  	\node at (plots c1r1.north east) [inner sep=0pt,anchor=north, yshift=10ex] {\ref{grouplegend2}};
  	%\node[rotate=90] at (plots c1r1.south west) [yshift=7ex] {Test accuracy};
  	%\node[rotate=-90] at (plots c1r1.south east) [yshift=54ex] {Train loss};
\end{tikzpicture}

\caption{Training of convolutional neural networks on the CIFAR-10 dataset. (a): 10 hidden dense layers, (b): 30 hidden layers. The batch size is 512. The schedules are $\gamma_n=1\e-3$ and $\sigma=2e-4$ for epochs 1 to 15 and $\gamma_n=1e-4$ and $\sigma=0$ beyond.}
\label{fig:conv_dense:1}
\end{figure}
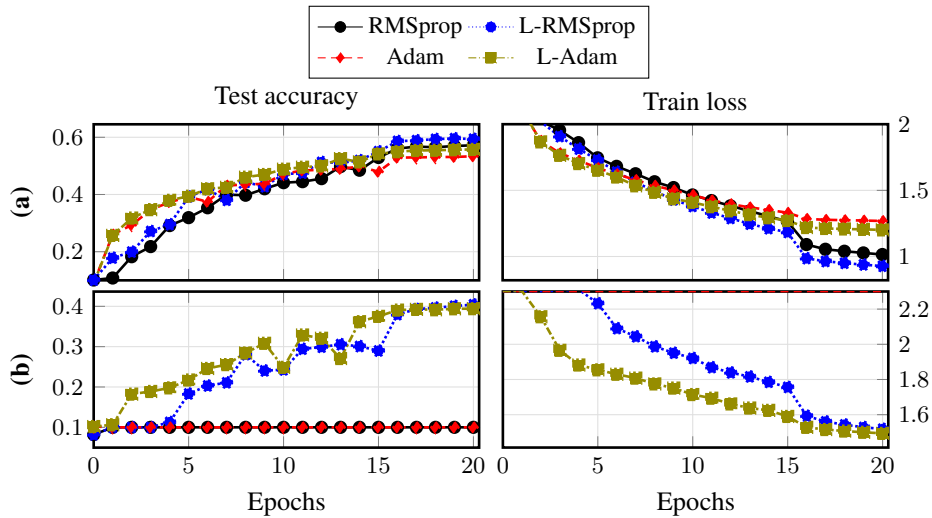

We then perform simulations in a similar setup on convolutional architectures that are more adapted to image recognition \cite{jarrett2009} followed by a large number of hidden dense layers.
More specifically, we train neural networks composed of two convolutional layers with $4 {\times} 4$ kernel size and 32 channels for each; $2 {\times} 2$ max-pooling is used after each convolutional layer. These layers are followed by respectively 10 and 30 hidden dense layers with 64 units each and by one dense output layer.
Since the images in the CIFAR-10 dataset do not have a good resolution, we cannot expect a very high
accuracy on the test set. Instead, we focus on comparing different algorithms on the same model architecture.
The results are given in Figure \ref{fig:conv_dense:1} and we make similar observations: Langevin algorithms show improvements with 10 hidden dense layers and for 30 dense layers, non-Langevin algorithms run into vanishing gradient issues which is not the case for Langevin algorithms.

\subsection{Highway networks}

We now perform the same simulations on Highway networks, in a setting very similar to \cite{highway}. Comparably to residual networks, the output of a highway layer is a convex combination of the output of a dense layer and the output of a identity layer; the parameter controlling the convex combination is itself trainable. For a layer with weights $(\theta_D, \theta_T)$, the output reads
\begin{equation}
y = T_{\theta_T}(x) \cdot D_{\theta_D}(x) + (1-T_{\theta_T}(x)) \cdot x,
\end{equation}
where $T$ and $D$ are dense layers and where $T$ has sigmoid output.

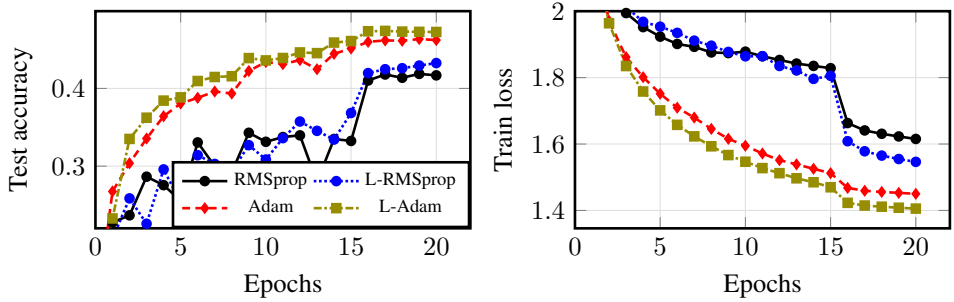
\begin{figure}
\centering

\begin{tikzpicture}
\begin{axis}[
	xlabel=Epochs,
	ylabel=Test accuracy,
	xmin=0,
	ymin=0.22,
	legend style={at={(1,0)},anchor=south east, font=\scriptsize},
	legend columns=2,
	grid=both,
	minor grid style={gray!25},
	major grid style={gray!25},
	width=0.47\linewidth,
	height=0.16\paperheight,
	%no marks,
	line width=1pt,
	mark size=1.5pt,
	mark options={solid},
	]
	\addplot[color=black, mark=*] %
		table[x=time,y=f,col sep=comma]{./data/highway/rmsprop_80x64/0_v_acc.csv};
	\addlegendentry[color=black]{RMSprop};
	\addplot[color=blue, mark=*, style=densely dotted] %
		table[x=time,y=f,col sep=comma]{./data/highway/rmsprop_80x64/1_v_acc.csv};
	\addlegendentry[color=black]{L-RMSprop};
	\addplot[color=red, mark=diamond*, style=densely dashed] %
		table[x=time,y=f,col sep=comma]{./data/highway/adam_80x64/0_v_acc.csv};
	\addlegendentry[color=black]{Adam};
	\addplot[color=olive, mark=square*, style=densely dashdotted] %
		table[x=time,y=f,col sep=comma]{./data/highway/adam_80x64/1_v_acc.csv};
	\addlegendentry[color=black]{L-Adam};
\end{axis}
\end{tikzpicture}%
~
\begin{tikzpicture}
\begin{axis}[
	xlabel=Epochs,
	ylabel=Train loss,
	xmin=0,
	ymax=2,
	legend style={at={(1,0)},anchor=south east,},
	grid=both,
	minor grid style={gray!25},
	major grid style={gray!25},
	width=0.47\linewidth,
	height=0.16\paperheight,
	%no marks,
	line width=1pt,
	mark size=1.5pt,
	mark options={solid},
	]
	\addplot[color=black, mark=*] %
		table[x=time,y=f,col sep=comma]{./data/highway/rmsprop_80x64/0_loss.csv};
	\addplot[color=blue, mark=*, style=densely dotted] %
		table[x=time,y=f,col sep=comma]{./data/highway/rmsprop_80x64/1_loss.csv};
	\addplot[color=red, mark=diamond*, style=densely dashed] %
		table[x=time,y=f,col sep=comma]{./data/highway/adam_80x64/0_loss.csv};
	\addplot[color=olive, mark=square*, style=densely dashdotted] %
		table[x=time,y=f,col sep=comma]{./data/highway/adam_80x64/1_loss.csv};
\end{axis}
\end{tikzpicture}

\caption{Training of a highway neural network with 80 dense hidden layers. The schedules are $\gamma_n=1\e-3$ and $\sigma=1e-4$ for epochs 1 to 15 and $\gamma_n=1e-4$ and $\sigma=0$ beyond.}
\label{fig:highway:1}
\end{figure}

We observe that Langevin algorithms become effectively faster than non-Langevin algorithms only from a larger depth than for plain networks.
In Figure \ref{fig:highway:1} we plot the results for the training of a network composed of 80 dense hidden layers with 64 units each and ReLU activation on the CIFAR-10 dataset, showing the possibilities of acceleration through Langevin algorithms, even in a residual (highway) architecture.
%We also observe that the gains brought by the Langevin algorithms vary depending on the choice of the preconditioner. They are indeed higher for the RMSprop preconditioner, and for the Adadelta preconditioner for very deep highway networks.

\section{Layer Langevin algorithm}
\label{sec:layer_langevin}

We introduce a new Langevin algorithm for stochastic optimization of deep neural networks that we call Layer Langevin algorithm.
Choosing a preconditioner rule $P$, some weights are updated following the Langevin rule while the other weights are updated following the non-Langevin rule. Denoting $\theta^{(i)}_n$ the $i$th weight at step $n$, we have for every $i$:
\begin{align}
\label{eq:layer_langevin}
\theta^{(i)}_{n+1} & = \theta^{(i)}_n - \gamma_{n+1} [P_{n+1} \cdot g_{n+1}]^{(i)} + \mathds{1}_{i \in \mathcal{J}} \sigma \sqrt{\gamma_{n+1}} \big[\mathcal{N}(0, P_{n+1}) \big]^{(i)},
\end{align}
where $\mathcal{J}$ is a subset of weight indices and where $P_n$ denotes the preconditioner.
To simplify the choice of $\mathcal{J}$, we choose $\mathcal{J}$ as the subset of indexes of weights belonging to some layers. However, a finer control over the subset $\mathcal{J}$ remains possible. To implement this method in practice, we simply assign before the training an attribute equals to $\mathds{1}_{i \in \mathcal{J}}$ to every trainable variable of the network.

\begin{figure}
\centering

\begin{tikzpicture}
\begin{axis}[
	xlabel=Epochs,
	ylabel=Test accuracy,
	xmin=0,
	ymin=0.22,
	legend style={at={(1,0)},anchor=south east, font=\scriptsize},
	legend columns=2,
	grid=both,
	minor grid style={gray!25},
	major grid style={gray!25},
	width=0.5\linewidth,
	height=0.2\paperheight,
	%no marks,
	line width=1pt,
	mark size=1.5pt,
	mark options={solid},
	]
	\addplot[color=black, mark=*] %
		table[x=time,y=f,col sep=comma]{./data/layer_langevin/adam_30x64/0_v_acc.csv};
	\addlegendentry[color=black]{Adam};
	\addplot[color=blue, mark=*, style=densely dotted] %
		table[x=time,y=f,col sep=comma]{./data/layer_langevin/adam_30x64/4_v_acc.csv};
	\addlegendentry[color=black]{L-Adam};
	\addplot[color=red, mark=diamond*, style=densely dashed] %
		table[x=time,y=f,col sep=comma]{./data/layer_langevin/adam_30x64/1_v_acc.csv};
	\addlegendentry[color=black]{LL-Adam 30 ${\%}$};
	\addplot[color=olive, mark=square*, style=densely dashdotted] %
		table[x=time,y=f,col sep=comma]{./data/layer_langevin/adam_30x64/2_v_acc.csv};
	\addlegendentry[color=black]{LL-Adam 60 ${\%}$};
	\addplot[color=Brown, mark=triangle*,]
		table[x=time,y=f,col sep=comma]{./data/layer_langevin/adam_30x64/3_v_acc.csv};
	\addlegendentry[color=black]{LL-Adam 90 ${\%}$};
\end{axis}
\end{tikzpicture}%
~
\begin{tikzpicture}
\begin{axis}[
	xlabel=Epochs,
	ylabel=Train loss,
	xmin=0,
	ymax=1.4,
	legend style={at={(1,0)},anchor=south east,},
	grid=both,
	minor grid style={gray!25},
	major grid style={gray!25},
	width=0.5\linewidth,
	height=0.2\paperheight,
	%no marks,
	line width=1pt,
	mark size=1.5pt,
	mark options={solid},
	]
	\addplot[color=black, mark=*] %
		table[x=time,y=f,col sep=comma]{./data/layer_langevin/adam_30x64/0_loss.csv};
	\addplot[color=blue, mark=*, style=densely dotted] %
		table[x=time,y=f,col sep=comma]{./data/layer_langevin/adam_30x64/4_loss.csv};
	\addplot[color=red, mark=diamond*, style=densely dashed] %
		table[x=time,y=f,col sep=comma]{./data/layer_langevin/adam_30x64/1_loss.csv};
	\addplot[color=olive, mark=square*, style=densely dashdotted] %
		table[x=time,y=f,col sep=comma]{./data/layer_langevin/adam_30x64/2_loss.csv};
	\addplot[color=Brown, mark=triangle*,]
		table[x=time,y=f,col sep=comma]{./data/layer_langevin/adam_30x64/3_loss.csv};
\end{axis}
\end{tikzpicture}

\caption{Layer Langevin method comparison on a dense neural network with 30 hidden layers. The schedules are $\gamma_n=1\e-3$ and $\sigma=5e-4$ for epochs 1 to 13 and $\gamma_n=1e-4$ and $\sigma=0$ beyond.}
\label{fig:layer_langevin:1}
\end{figure}
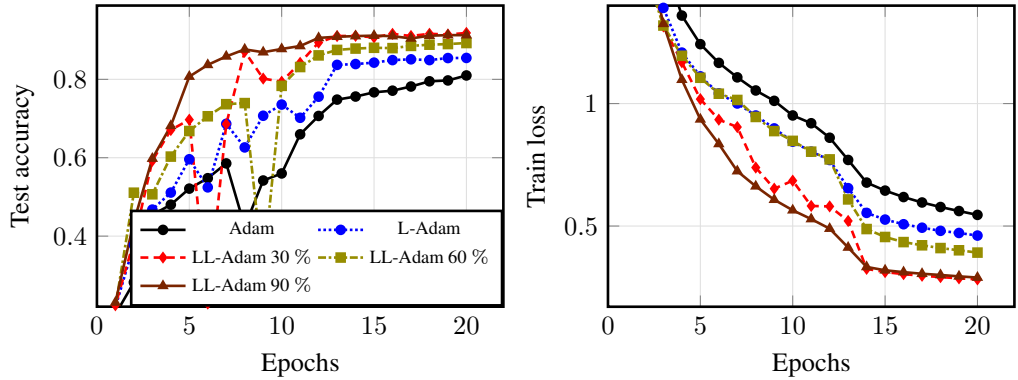

We compare the performances of Layer Langevin algorithms with the Adam preconditioner for different choices of the subset of layers. The results are given in Figure \ref{fig:layer_langevin:1} for the training of a dense network with 30 hidden dense layers on the MNIST dataset in a setting similar to Figure \ref{fig:dense:1}. For some optimizer \textit{Name}, we denote LL-\textit{Name} $p\%$ the corresponding Layer Langevin algorithm where the subset $\mathcal{J}$ is the first $p\%$ layers of the network. We observe that we obtain significant gains in comparison with the vanilla Langevin algorithm and that the best performances are obtained when choosing the subset $\mathcal{J}$ as being the first $\ell$ layers for some $\ell \in \mathbb{N}$, in particular all the layers of the network except the few last ones.

%
%We then test the Layer Langevin algorithm where we apply the Langevin update on the first $L$ layers of the network for different values of $L$. The results are given in Figure \ref{fig:CIFAR_layer_L} for the RMSprop precondtioner.
%We observe empirically that the best performances are obtained when the Langevin update is applied not to all the layers but to all the layers except the last few ones. In our case, we set $L=18$ for a 21-layer deep network.
%
%We also experiment the Layer Langevin algorithm on highway networks. The results are given in Figure \ref{fig:highway_layer_L}. Similarly, the best performances are obtained in the case where the Langevin update is applied to all the layers except the few last ones (for $L=72$ for a 82-layer deep neural network).
%
%These experiments show that Layer Langevin algorithm is an improvement of preconditioned Langevin algorithms for the training of very deep neural networks, including deep highway networks.
%Layer Langevin algorithms allow to escape from traps that essentially result from the nonlinearities of the first layers, and do not add noise to the last layers which may not need it.
%

\section{Application to deep architectures for image recognition}

We now test the Layer Langevin algorithm to speed up the training of neural networks with very deep architectures that are popular in image recognition.
VGG (Visual Geometry Group) network \cite{simonyan2015} consists in a large number of successive 2D convolutional layers with ReLU activation; the size of the image is gradually reduced using $2 {\times} 2$ pooling layers. However its performances are limited by the difficulty of training very deep networks. To cope with this issue, residual network (ResNet) \cite{residual} adds residual connections to the VGG architecture. For $H_\ell$ some layer composed of convolutions, activations and batch normalizations, the output is $x_{\ell+1} = H_\ell(x_\ell) + x_\ell$ instead of simply $H_\ell(x_\ell)$, so that the residual layer behaves in part as the identity layer. Similarly to highway networks, residual connections improve the flow of gradient inside the network.
%Instead of summing which implies a loss of information, in dense convolutional network (DenseNet) \cite{huang2017} the output $x_{\ell+1}$ depends on the concatenation of the $r$ preceding outputs i.e. is given by $H_\ell([x_{\ell-r+1}, \ldots, x_\ell])$.

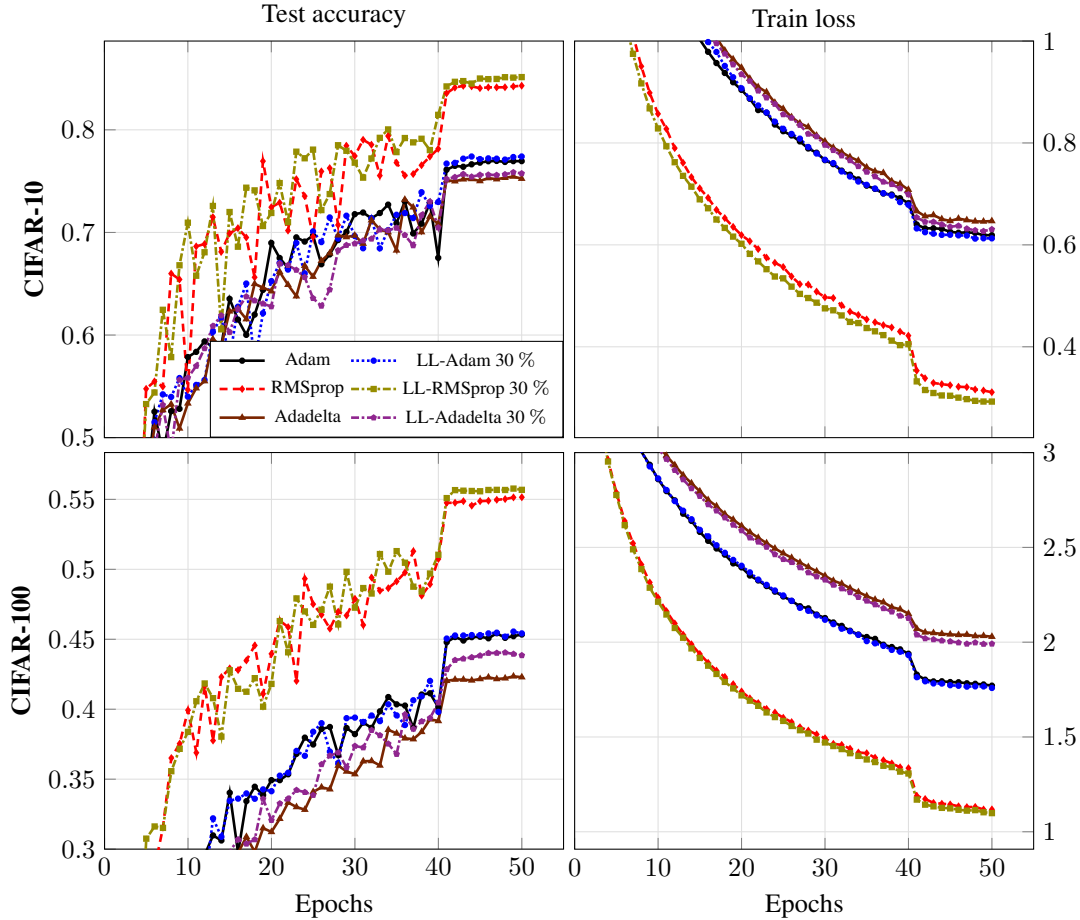
\begin{figure}
\centering

\begin{tikzpicture}
	\begin{groupplot}[group style={
					group size=2 by 2,
					horizontal sep=0.15cm,
					vertical sep=0.2cm,
					x descriptions at=edge bottom,
					group name=plots,
					},
			  ytick pos=left,
			  height=0.3\textheight,
			  width=0.55\textwidth,
			  every axis plot/.append style={line width=1pt,mark size=0.7pt,mark options={solid}},
			  xmin=0,
			  xlabel=Epochs,
			  grid=both,
			  legend style={at={(1,0)},anchor=south east, font=\scriptsize},
			  legend columns=2,
			  minor grid style={gray!25},
			  major grid style={gray!25},
			  ]

  		\nextgroupplot[ymin=0.5,title=Test accuracy, title style={yshift=-1ex}, ylabel=\textbf{CIFAR-10}]
	\addplot[color=black, mark=*]
		table[x=time,y=f,col sep=comma]{./data/resnet/ladam_2x5_cifar10/0_v_acc.csv};
	\addlegendentry[color=black]{Adam};
	\addplot[color=blue, mark=*, style=densely dotted]
		table[x=time,y=f,col sep=comma]{./data/resnet/ladam_2x5_cifar10/1_v_acc.csv};
	\addlegendentry[color=black]{LL-Adam 30 ${\%}$};
	\addplot[color=red, mark=diamond*, style=densely dashed]
		table[x=time,y=f,col sep=comma]{./data/resnet/lrmsprop_2x5_cifar10/0_v_acc.csv};
	\addlegendentry[color=black]{RMSprop};
	\addplot[color=olive, mark=square*, style=densely dashdotted]
		table[x=time,y=f,col sep=comma]{./data/resnet/lrmsprop_2x5_cifar10/1_v_acc.csv};
	\addlegendentry[color=black]{LL-RMSprop 30 ${\%}$};
	\addplot[color=Brown, mark=triangle*,]
		table[x=time,y=f,col sep=comma]{./data/resnet/ladadelta_2x5_cifar10/0_v_acc.csv};
	\addlegendentry[color=black]{Adadelta};
	\addplot[color=Plum, mark=pentagon*, style=densely dashdotted]
		table[x=time,y=f,col sep=comma]{./data/resnet/ladadelta_2x5_cifar10/1_v_acc.csv};
	\addlegendentry[color=black]{LL-Adadelta 30 ${\%}$};

		\nextgroupplot[ymax=1, title=Train loss, title style={yshift=-1ex}, ytick pos=right]
	\addplot[color=black, mark=*]
		table[x=time,y=f,col sep=comma]{./data/resnet/ladam_2x5_cifar10/0_loss.csv};
	\addplot[color=blue, mark=*, style=densely dotted]
		table[x=time,y=f,col sep=comma]{./data/resnet/ladam_2x5_cifar10/1_loss.csv};
	\addplot[color=red, mark=diamond*, style=densely dashed]
		table[x=time,y=f,col sep=comma]{./data/resnet/lrmsprop_2x5_cifar10/0_loss.csv};
	\addplot[color=olive, mark=square*, style=densely dashdotted]
		table[x=time,y=f,col sep=comma]{./data/resnet/lrmsprop_2x5_cifar10/1_loss.csv};
	\addplot[color=Brown, mark=triangle*,]
		table[x=time,y=f,col sep=comma]{./data/resnet/ladadelta_2x5_cifar10/0_loss.csv};
	\addplot[color=Plum, mark=pentagon*, style=densely dashdotted]
		table[x=time,y=f,col sep=comma]{./data/resnet/ladadelta_2x5_cifar10/1_loss.csv};

  		\nextgroupplot[ymin=0.3, ylabel=\textbf{CIFAR-100}]
	\addplot[color=black, mark=*]
		table[x=time,y=f,col sep=comma]{./data/resnet/ladam_2x5_cifar100/0_v_acc.csv};
	\addplot[color=blue, mark=*, style=densely dotted]
		table[x=time,y=f,col sep=comma]{./data/resnet/ladam_2x5_cifar100/1_v_acc.csv};
	\addplot[color=red, mark=diamond*, style=densely dashed]
		table[x=time,y=f,col sep=comma]{./data/resnet/lrmsprop_2x5_cifar100/0_v_acc.csv};
	\addplot[color=olive, mark=square*, style=densely dashdotted]
		table[x=time,y=f,col sep=comma]{./data/resnet/lrmsprop_2x5_cifar100/1_v_acc.csv};
	\addplot[color=Brown, mark=triangle*,]
		table[x=time,y=f,col sep=comma]{./data/resnet/ladadelta_2x5_cifar100/0_v_acc.csv};
	\addplot[color=Plum, mark=pentagon*, style=densely dashdotted]
		table[x=time,y=f,col sep=comma]{./data/resnet/ladadelta_2x5_cifar100/1_v_acc.csv};

		\nextgroupplot[ymax=3, ytick pos=right]
	\addplot[color=black, mark=*]
		table[x=time,y=f,col sep=comma]{./data/resnet/ladam_2x5_cifar100/0_loss.csv};
	\addplot[color=blue, mark=*, style=densely dotted]
		table[x=time,y=f,col sep=comma]{./data/resnet/ladam_2x5_cifar100/1_loss.csv};
	\addplot[color=red, mark=diamond*, style=densely dashed]
		table[x=time,y=f,col sep=comma]{./data/resnet/lrmsprop_2x5_cifar100/0_loss.csv};
	\addplot[color=olive, mark=square*, style=densely dashdotted]
		table[x=time,y=f,col sep=comma]{./data/resnet/lrmsprop_2x5_cifar100/1_loss.csv};
	\addplot[color=Brown, mark=triangle*,]
		table[x=time,y=f,col sep=comma]{./data/resnet/ladadelta_2x5_cifar100/0_loss.csv};
	\addplot[color=Plum, mark=pentagon*, style=densely dashdotted]
		table[x=time,y=f,col sep=comma]{./data/resnet/ladadelta_2x5_cifar100/1_loss.csv};

\end{groupplot}
\end{tikzpicture}

\caption{Layer Langevin method comparison for the training of ResNet-20. The initial number of channels is 16. The schedules are $\gamma_n=1\e-3$ ($2\e-1$ for Adadelta) and $\sigma=5\e-4$ ($5\e-3$ for Adadelta) for epochs 1 to 40 $\gamma_n$ is divided by 10 and $\sigma$ is set to 0 beyond.}
\label{fig:resnet:1}
\end{figure}

We train on the CIFAR-10 dataset a ResNet architecture composed of 2 blocks with 5 residual layers each; each block is followed by a size reduction layer. This architecture is given as ResNet-20 in \cite[Section 4.2]{residual}.
%We train the DenseNet121 architecture \cite{huang2017} on the CIFAR-100 dataset and we plot the results in Figure \ref{fig:densenet:1}.
We apply usual data augmentation to both CIFAR-10 and CIFAR-100 datasets \cite{lee2015,residual}: 4 pixels are padded on each side and a $32 \times 32$ crop is randomly sampled from the padded image or its horizontal flip.
The results are given in figure \ref{fig:resnet:1}.
Experiments show that Layer Langevin algorithms (in this case LL-Adam 30${\%}$) yield improvements in comparison with non-Langevin methods, even on residual architectures adapted to very deep learning. The train loss is also plotted, showing that the better performances of Layer Langevin is not only due to some overfitting effect.

\begin{table}
\centering
\begin{tabular}{ccccccc}
\hline
& Adam & LL-Adam & RMSprop & LL-RMSprop & Adadelta & LL-Adadelta \\ 
\hline 
CIFAR-10 & 76.95 \% & 77.39 \% & 84.29 \% & 85.14 \% & 75.23 \% & 75.74 \%  \\
\hline
CIFAR-100 & 45.33 \% & 45.41 \% & 55.15 \% & 55.68 \% & 42.28 \% & 43.84 \% \\
\hline
\end{tabular}
\caption{Final test accuracy values obtained in Figures \ref{fig:resnet:1}.}
\label{table:resnet}
\end{table}

%
%\section*{Discussion}
%
%Let us now discuss of the limitations of our article.
%
%%The Langevin algorithms that are discussed and introduced are advantageous only for very deep neural networks whereas in practice, for many applications, neural networks do not need that many layers.
%
%We are aware that the training of very deep neural networks is a problem that has been extensively studied. That is why we paid a particular attention to test our algorithms on highway networks, which were introduced specifically to deal with this issue.
%
%Using such Langevin algorithms requires adjusting the noise parameter $\sigma \in (0,\infty)$, problem which is similar to the adjustment of the learning rate for all optimization algorithms.
%
%As noted in Section \ref{sec:layer_langevin}, a finer control on the choice of $\mathcal{J}$ in \eqref{eq:layer_langevin} remains possible, and most of our results on the Layer Langevin algorithm are empirical. A theoretical approach would help improve further our algorithm but it seems complex to us.
%
%
%
\begin{ack}
I would like to thank Gilles Pag\`es for insightful discussions.
\end{ack}
\small

%\bibliographystyle{alpha}
%\bibliography{../../biblio_all}

\newcommand{\etalchar}[1]{$^{#1}$}

\normalsize

\end{document}